\documentclass{article}
\pdfoutput=1

\usepackage[top=0.85in,left=2.75in,footskip=0.75in]{geometry}
\usepackage{url}

\usepackage{changepage}
\usepackage{subfig}
\usepackage[utf8x]{inputenc}

\usepackage{textcomp,marvosym}

\usepackage{cite}

\usepackage{nameref,hyperref}
\usepackage[T1]{fontenc}

\usepackage{microtype}
\DisableLigatures[f]{encoding = *, family = * }

\usepackage[table]{xcolor}
\usepackage{booktabs} 

\newcolumntype{+}{!{\vrule width 2pt}}

\usepackage[aboveskip=1pt,labelfont=bf,labelsep=period,justification=raggedright,singlelinecheck=off]{caption}

\makeatletter
\renewcommand{\@biblabel}[1]{\quad#1.}
\makeatother

\usepackage{lastpage,fancyhdr,graphicx}
\usepackage{epstopdf}
\usepackage[normalem]{ulem}
 \usepackage{lscape}

\usepackage{textcomp,marvosym}

\usepackage{cite}

\usepackage{nameref,hyperref}

\usepackage{microtype}
\DisableLigatures[f]{encoding = *, family = * }

\usepackage{graphicx,wrapfig,lipsum}

\def\BibTeX{{\rm B\kern-.05em{\sc i\kern-.025em b}\kern-.08em
    T\kern-.1667em\lower.7ex\hbox{E}\kern-.125emX}}

\usepackage{cite}
\usepackage{amsmath,amssymb,amsfonts}
\usepackage{algorithmic}
\usepackage{graphicx}
\usepackage{textcomp}

\usepackage{times}  
\usepackage{helvet}  
\usepackage{courier}  
\usepackage{url}  

\usepackage{blindtext, graphicx}
\usepackage[english]{babel}
\usepackage{amsmath}
\usepackage{mathtools}
\usepackage{graphicx}

\usepackage{pgffor}
\usepackage{enumerate}
\usepackage{enumitem} 
\usepackage{chemgreek,textgreek}
\usepackage{tikz}
\usepackage{pifont}
\usepackage{ulem}
\useunder{\uline}{\ul}{}
\usepackage{array}
\usepackage{float}
\usepackage{multirow}                

\newcommand{\cmark}{\ding{51}}%
\newcommand{\xmark}{\ding{55}}%

\usepackage{graphicx,wrapfig,lipsum}

\usepackage{color, colortbl}
\definecolor{Gray}{gray}{0.9}
\definecolor{Cyan}{RGB}{204, 255, 255}
\definecolor{Yellow}{RGB}{255, 242, 204}
\definecolor{Red}{RGB}{255, 204, 204}
\newcolumntype{a}{>{\columncolor{Gray}} c}

\usepackage{authblk}

\begin{document}

\title{Analyzing and learning the language for different types of harassment} 

\author[3+]{Mohammadreza Rezvan}
\author[1+]{Saeedeh Shekarpour}
\author[2]{Faisal Alshargi}
\author[3]{Krishnaprasad Thirunarayan}
\author[3]{Valerie L. Shalin}
\author[3]{Amit Sheth}

\affil[1]{Email: sshekarpour1@udayton.edu, University of Dayton, Dayton, USA}
\affil[2]{Email: alshargi@informatik.uni-leipzig.de, University of Leipzig, Germany}
\affil[3]{ Knoesis Center, Wright State University, Dayton, USA}
\affil[+]{ These authors contributed equally to this work.}




\maketitle

\section{Abstract}

\textbf{Disclaimer:} This paper is concerned with violent online harassment. To describe the subject at an adequate level of realism, examples of our collected tweets involve violent, threatening, vulgar and hateful speech language in the context of racial, sexual, political, appearance and intellectual harassment. 
While these examples are shared to portray reality, readers are alerted in advance and may wish to avoid reading this material if it could cause discomfort and disagreeable response.

The presence of a significant amount of harassment in user-generated content and its negative impact calls for robust automatic detection approaches. This requires the identification of different types of harassment. Earlier work has classified harassing language in terms of hurtfulness, abusiveness, sentiment, and profanity. However, to identify and understand harassment more accurately, it is essential to determine the contextual type that captures the interrelated conditions
in which harassing language occurs.
In this paper


we introduce the notion of contextual type in harassment by distinguishing between five contextual types: (i) sexual, (ii) racial, (iii) appearance-related, (iv) intellectual and (v) political.
We utilize an annotated corpus from Twitter distinguishing these types of harassment. We study the context of each kind to shed light on the linguistic meaning, interpretation, and distribution, with results from two lines of investigation: an extensive linguistic analysis, and the statistical distribution of uni-grams. 
We then build type- aware classifiers to automate the identification of type-specific harassment.
Our experiments demonstrate that these classifiers provide competitive accuracy for identifying and analyzing harassment on social media.
We present extensive discussion and significant observations about the effectiveness of type-aware classifiers using a detailed comparison setup, providing insight into the role of type-dependent features.



\section{Introduction}
\textbf{Disclaimer:} This paper is concerned with violent online harassment. To describe the subject at an adequate level of realism, examples of our collected tweets involve violent, threatening, vulgar and hateful speech language in the context of racial, sexual, political, appearance and intellectual harassment. 
While these examples are shared to portray reality, readers are alerted in advance and may wish to avoid reading this material if it could cause discomfort and disagreeable response.

Although social media has enabled people to connect and interact with each other, it has also made people vulnerable to insults, humiliation, hate, bullying--facing threats from individuals who are either known (e.g., colleagues, friends) or unknown (e.g., fans, clients, anonymous entities).
A Pew research center report  \cite{Foot12} (e.g., offensive name-calling, and shaming, In this work, cyberbullying and harassment are used interchangeably.). One-in-five (18\%) victims characterized their exposure as severe. The resulting negative impact from emotional distress, privacy concerns and threats to physical safety and mental health, affect individuals online and offline. 
This calls for tool-based, automatic detection, monitoring, and analysis of hurtful language to protect online users. 
The prior state-of-the-art is limited to detecting specific hurtful language such as hate speech \cite{cite31}, abusive language \cite{cite30}, and profanity \cite{cite32}, collectively termed Negative Affective Language (NAL).
In the following, we present the definitions and terms for variants of harassing language:

\begin{itemize}
    \item \textbf{Hate speech} is ``speech that denigrates a person because of their innate and protected characteristics'' \cite{ElSherief2018HateLA}. Furthermore, it is divided into two categories: \emph{directed} and \emph{generalized}, depending upon whether there is an explicit target or not. 
    
   \item \textbf{Abusive Language} is  ``the collection and misuse of private user information, cyberbullying and the distribution of offensive, misleading, false or malicious information''\cite{abusivelanguage}.  
    
   \item \textbf{Offensive Language} employs profanity, is strongly impolite, rude or vulgar red {expressed} with fighting or hurtful words to insult a targeted individual or group \cite{cite34, cite36,cite2,abusivebehavious}.  
    
 \item \textbf{Aggressive Language} shows overt, angry and often violent social interaction with the intention of inflicting damage or other unpleasantness upon another individual or group of people \cite{cite35,cite5}.     
    
    \item \textbf{Harassing (Cyberbullying) Language} is the use of force, threat, or coercion to abuse, embarrass, intimidate, or aggressively dominate others. It typically denotes repeated and hostile behavior performed by a group or an individual \cite{cite35,cite1,cite5}.

\end{itemize}

These definitions are highly subjective and overlap, making them hard to differentiate.  For example, the definition of harassing language is similar to aggressive language. 
We posit that all of these NALs are \textbf{hurtful} and thus \textbf{harassing.} But they might vary in their level of severity, presence or absence of target (victim), contextual interpretation and purpose. 
In this paper, we frame harassing language as  \emph{offensive language where a given post/message contains ``profanity, strongly impolite, rude, vulgar or threatening language''}.


\begin{wrapfigure}{r}{0.55\textwidth}
\centering

\includegraphics[width=0.55\columnwidth]{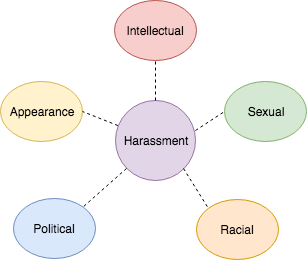}

\caption{Five contextual types of harassment.}
\label{fig:type}
\end{wrapfigure}
State-of-the-art harassment detection fails to exploit the \textbf{contextual type} of harassing language.
Webster's dictionary \cite{Foot13} provides the following definition for context: ``the parts of a discourse that surround a word or passage and can throw light on its meaning''. 
Here, we describe the notion of contextual type as the linguistic or statistical conditions that help in differentiating the type of harassment.
For example, the circumstance of a student who has been subjected to sexual harassment by her ex-partner differs from a student racially harassed because of her/his color.
We suggest that \emph{contextual type} influences the linguistic characteristics of harassment.
We propose five contextual types of harassment in online communication on social media: (i) sexual harassment, (ii) racial harassment, (iii) appearance-related harassment, (iv) intellectual harassment, and (v) political harassment. This categorization is represented in Fig. \ref{fig:type}.
Below, we define each type of harassment using illustrative examples from the Twitter corpus we have created.

\begin{enumerate}
    
\item \emph{Sexual harassment} is  offensive sexual speech that usually targets females. E.g., the harasser might comment on the victim's body in a vulgar manner or mention sexual relationships in an aggressive way. 
Note that using sexually profane words is not sufficient to indicate offensive sexual harassment \cite{Foot5,Foot6}.


\item \emph{Racial harassment} targets race and ethnicity characteristics of a victim such as skin color, country of origin, culture, or religion, in an offensive manner\cite{Foot7}. 


\item \emph{Appearance-related harassment} uses embarrassing language referring to body appearance. Fat shaming \cite{cite15} and body shaming are key subtypes of this type of harassment.


\item \emph{Intellectual harassment} offends the intellectual power or opinions of individuals.
Even smart people may be ridiculed and become victims\cite{Foot8}. 

\item \emph {Political harassment} is related to someone's political views\cite{Foot9}.
Typical targets are politicians and politically inclined individuals who receive threatening messages \cite{Foot10}.


\end{enumerate}

Determining the real intent behind a tweet regarding the type of harassment can have serious implications for public perception. 
Consider the controversial tweet from Roseanne Barr targeting Valerie 
Jarett:
characterizing Jarrett, an African-American woman born in Iran, as a child of the Muslim Brotherhood and an ape\cite{Foot11}. Twitter Users regarded this tweet as racist, while Barr defended herself as making a bad joke about Jarrett's politics and looks. Thus, whether the tweet is considered to be racist or regarded as appearance-related or political makes a significant difference.
Reliable assessment of the type of harassment can have significant repercussions.
We are unaware of any prior work on studying harassment concerning these five types.

We summarize our contributions as follows:   
(i) We introduce five contextual types of harassment. Then, we provide a systematic, and comparative analysis to assess offensive language from linguistic and statistical perspectives for each contextual type. This allows us to exploit relevant features for developing classifiers to identify these critical types of harassment on social media.
(ii) We develop type-aware classifiers and capture their effectiveness using a detailed comparative study.
This paper is organized as follows. The next section reviews the related literature.
We then present the type-aware corpus that we have developed.
Subsequently, we analyze our compiled corpus linguistically as well as statistically, which shows us the significant type-specific features for various types of harassment. We then discuss supervised learning approaches and classifiers for detecting the harassing language in comparative settings. We also provide an error analysis study regarding the pitfalls and challenges of our strategy.
We close with the conclusions and our future plans.

\section*{State-of-the-art in Harassment Research}
\label{sec:relatedwork}


The previous research studies targeted various social media sources such as Twitter, Instagram, and Facebook. In Table \ref{table:RelatedWorkSummery}, we summarize the prior literature with their corresponding goals, conclusions, and underlying data sets.
Here, we specifically note particularly prominent related work.
In \cite{cite1}, the authors seek to predict cyberbullying incidents on Instagram.  They built a predictive model for the incidence of cyberbullying using features from initially posted data, a social graph, and temporal properties. 
The work in \cite{cite18} proposed an approach for detecting harassment features based on the content, sentiment, and context. Using Slashdot and MySpace data, they showed significant improvement using TFIDF supplemented with sentiment and contextual features.
The authors of \cite{cite16} proposed an approach to spotting harassers as well as victims on social media. They considered the social structure and infer which user is a likely instigator and which user is expected to be a victim. This model is based on social interactions and the language of users in social media.
Similarly, \cite{cite17} proposes a method that simultaneously discovers instigators and victims of bullying incidents. 
It extends an initial bullying vocabulary using twitter and ask.fm. 
In \cite{cite19}, the authors proposed a supervised learning method for detecting cyberbullying in Japan. 
In \cite{cite8}, the authors propose a supervised learning method based on \textit{fuzzy logic} and \textit{genetic algorithm} to identify the presence of cyberbullying terms and classify activities, such as flaming, harassment, racism, and terrorism on social media. Fuzzy rules were used to classify data, and a genetic algorithm was used for optimizing the parameters.

\cite{cite2} explores the correlation of behaviors and actions of people and their emotions. The authors developed a large emotion-labeled dataset of harassing tweets. They applied 131 emotion hashtag keywords categorized into seven groups and collected 5 million tweets. 
 To find useful features for emotion identification, they applied LIBLINEAR \cite{cite3} and Multinomial Naive Bayes \cite{cite4} algorithms. They extracted n-gram features \cite{cite7} to analyze the emotion, and they applied \textit{Linguistic Inquiry and Word Count (LIWC)} to expand the feature set with related emotional words.
Interestingly, the authors of \cite{cite5} target cyber-aggression and cyberbullying in a multi-modal context with text comments and media objects on Instagram.
They concluded that non-text features are not able to substantially improve the performance of cyberbullying detection compared to text-based features.

Different from the previous work, 
some literature examines the \textbf{psychological implications} of harassment incidents \cite{cite27}. The authors in \cite{cite9} sought the reasons behind the updates of posts on Facebook. They noticed that: (i) the majority of posts are about social activities and everyday life, (ii) people with low self-esteem updated their status on relationship whereas those with high self-esteem update their status with respect to their children. Moreover, people with narcissistic personality disorder updated their status through their achievements. 
Furthermore, they observed a correlation between the number of likes and comments with esteem level of people (e.g., the people with the low self-esteem receive fewer likes and comments because their status expresses greater negative affect).
Similarly, the authors of \cite{cite10} discuss narcissism personality disorder in Facebook users and its implications in harassing incidents.
Our own past work \cite{venkatThesis,rajeshwariThesis} focused on (i) using a conversation between a sender and a receiver
to better capture its normal linguistic nature (e.g., base rates for curse word usage) and the nature 
of the relationship between participants (e.g., friends vs. strangers), and  
(ii) analyze comments/review threads to better identify offensive content in non-text media such as  YouTube videos \cite{rajeshwariThesis}, to reliably detect harassment between participants.

 \begin{landscape}

\begin{table*}[hpbt]
\centering
\begin {scriptsize}
\begin{tabular}{|l|l|l|l|}
\toprule
 \rowcolor{Gray}  \multicolumn{1}{|c|}{\textbf{Paper}} & \multicolumn{1}{c|}{\textbf{Goal}}                                                                                                         & \multicolumn{1}{c|}{\textbf{Data}}                                                                                                                                                                                                                                                                         & \multicolumn{1}{c|}{\textbf{Conclusions}}                                                                                                                                                                                                                    \\ \midrule
\cite{cite34}       & Detecting offensive and hateful speech language                                                                                            & \begin{tabular}[c]{@{}l@{}}85.4 Million Tweets Collected from 33458 twitter user using profane words. \\ 25000 tweets are selected.\end{tabular}                                                                                                      & \begin{tabular}[c]{@{}l@{}}Collected discriminating terms for \\ hate speech and offensive language\end{tabular}                                                                                                                                                      \\ \hline
\cite{cite35}       & \begin{tabular}[c]{@{}l@{}}Detecting aggressors  \\ and their behavior on social media\end{tabular}                                    & \begin{tabular}[c]{@{}l@{}}1.6 million tweets collected in 3 months, using crowd sourcing for annotation. \end{tabular}                                                                                                                                                          & \begin{tabular}[c]{@{}l@{}}Determined that posts of aggressor profiles \\ are more negative\end{tabular}                                                                                                                                                                   \\ \hline
\cite{cite36}       & \begin{tabular}[c]{@{}l@{}}Detecting offensive language \\ and identifying its sender.\end{tabular}                                         &    \begin{tabular}[c]{@{}l@{}}The data set includes comments from 2,175,474 Youtube users in \\reaction to the top 18 videos on different Topics.\end{tabular}                                                                                                                                                                                                                                                                                                   & \begin{tabular}[c]{@{}l@{}}(i) Conceptualized offensive content, and \\ (ii) enhanced features using lexical, \\     style, structural,      and context-specific features.\end{tabular}                                       \\ \hline
\cite{cite1}        & \begin{tabular}[c]{@{}l@{}}Predicting cyberbullying incidents on\\ Instagram social media\end{tabular}                                     & \begin{tabular}[c]{@{}l@{}} 41K users that are cyber bullied according to the random seed nodes. \\ 3165K tweets collected from 25K public users while \\ 697K Tweets labeled as profane tweets \end{tabular} & \begin{tabular}[c]{@{}l@{}}Classifier designed, trained, and applied for collecting data. \\  Logistic regression classifier \end{tabular}                     \\ \hline
\cite{cite18}      & \begin{tabular}[c]{@{}l@{}}Detecting harassment based wrt. content, sentiment, and context\end{tabular}                               & \begin{tabular}[c]{@{}l@{}}$\sim$11K tweets used in experiments \\ Fundacio'n Barcelona Media (FBM): Kongregate, Slashdot and MySpace. \\ Totally 10,951 tweets collected and nearly 167 labeled offensive.\end{tabular}                                              & \begin{tabular}[c]{@{}l@{}}Improving accuracy in detecting harassing language using \\ discussion-style and chat-style language\end{tabular}                                                                                                                 \\ \hline
\cite{cite16}       & \begin{tabular}[c]{@{}l@{}}Detecting harassers and victims \\ in cyberbullying incidents\end{tabular}                                      & \begin{tabular}[c]{@{}l@{}}Collected twitter data using profane words \\ Twitter data contains 180,355 users and 296,308 tweets.\end{tabular}                                                                                     & Accuracy improved wrt. network features.                                                                                                                                                                                                          \\ \hline
\cite{cite17}       & \begin{tabular}[c]{@{}l@{}} Detecting instigators and victims of bullying\end{tabular}        & \begin{tabular}[c]{@{}l@{}}180K profile on Twitter and $\sim$300K tweets using profane words as seed \end{tabular}                                                                                                                                                                                        & scoring level of cyber bully and victim.                                                                                                                                                                                             \\ \hline
\cite{cite19}      & Detecting cyber bullying in the Japanese community.                                                                                              & Data from Japanese secondary schools                                                                                                                                                                                                                                                                       & \begin{tabular}[c]{@{}l@{}}Automatically extract new vulgarities from the Internet \\ to keep their offensive lexicon up to date.\end{tabular}                                                                                                               \\ \hline
\cite{cite2}        & \begin{tabular}[c]{@{}l@{}}Understanding behavior and actions of \\ individuals using emotion detection\end{tabular}  & $\sim$2.5M tweets                                                                                                                                                                                                                                                                                          & \begin{tabular}[c]{@{}l@{}}tweets dataset using harassment-related and emotion hashtags \end{tabular}                                                                                                             \\ \hline
\cite{cite6}       & Detecting bullying incident on social networks                                                                                            & $\sim$2M tweets collected in 4 weeks                                                                                                                                                                                                                                                                       & \begin{tabular}[c]{@{}l@{}}Developed a practical method of text mining, clustering, \\ dimensionality reduction and classification.\end{tabular}                                                                          \\ \hline
\cite{cite8}        & \begin{tabular}[c]{@{}l@{}}Classifying cyberbullying activities \\ on social network\end{tabular}                                          & \begin{tabular}[c]{@{}l@{}}Collected data from 18,554 users data from \\ Formspring.Me and MySpace.\end{tabular}                                                                                                                                                                                                 & \begin{tabular}[c]{@{}l@{}} predicting cyber bullying using fuzzy logic \end{tabular} \\ \hline
\cite{cite9}        & \begin{tabular}[c]{@{}l@{}}investigating the correlation of  harassment on Facebook\end{tabular} & \begin{tabular}[c]{@{}l@{}}555 Facebook users in the United States \\ (59\% female; Mage = 30.90, SDage = 9.19)\end{tabular}                                                                                                                                 & \begin{tabular}[c]{@{}l@{}}Results show most of the updating posts \\ related to the intellect people, children, and who they are \\in the romantic relation.\end{tabular}  
\\ \hline
\cite{cite10}       & \begin{tabular}[c]{@{}l@{}}Identifying narcissism, activities on Facebook social media\end{tabular}                                      & 256 Facebook users from  locations around the world.                                                                                                                                                                                                                                              & \begin{tabular}[c]{@{}l@{}}Text mining for narcissistic using on the comment likes

\end{tabular}

\\ \hline
\cite{cite39}       & \begin{tabular}[c]{@{}l@{}}Automatic cyber bulling detection\\ on  social media text \end{tabular}      & English and Dutch corpora from \textit{ASKfm}social media sites.                                               & \begin{tabular}[c]{@{}l@{}} detecting signals of cyber \\ bulling on social media, about bullies, victims, and bystanders.

\end{tabular}
\\ \hline

\cite{cite40}       & \begin{tabular}[c]{@{}l@{}}Decompose the overall detection problem\\ into detection of sensitive topics,\\ lending itself into text classification. \end{tabular}  & corpus contain  4500  YouTube comments.                                                                                                                                                                                                                                              & \begin{tabular}[c]{@{}l@{}} Concluded binary classifier for individual \\labels outperform multiclass classifier.

\end{tabular}
\\ \hline

\cite{cite42}& \begin{tabular}[c]{@{}l@{}} Cyberbulling detection with in multi modal content.\end{tabular}& $\sim$K entries from Instagram and Vine Dataset& \begin{tabular}[c]{@{}l@{}} proposed cyberbulling detection framwork \\ \textbf{XBully} based on network representation leaning. 

\end{tabular}
\\ \hline
\cite{cite41}& \begin{tabular}[c]{@{}l@{}}Identification of fake content in online news.\end{tabular}& 980 entries from fakeNewsAMT and celebrity Dataset& \begin{tabular}[c]{@{}l@{}} Linguistic analysis shows the importance \\ of the lexical, syntactic, and semantic of content. 

\end{tabular}

\\ \bottomrule

\end{tabular}
\end {scriptsize}

\caption {\bf Summery of the related research.}
\label {table:RelatedWorkSummery}

\end{table*}
\end{landscape}

\section*{Type-aware Harassment Corpus}
We published a type-aware annotated corpus and lexicon in \cite{cite26}.
Our corpus consists of 25,000 annotated tweets for the five types of harassment content and is available on the Git repository\cite{Foot14}. In the following, we discuss our strategies for corpus compilation and annotation.
The identification of cyberbullying typically begins with a lexicon of potentially profane or offensive words.
We created a lexicon (compiled from online resources \cite{Foot15}\cite{Foot16} \cite{Foot17}\cite{Foot18} \cite{Foot19}) containing offensive words covering five different types of harassment context. 
The resulting compiled lexicon includes six categories: (i) sexual, (ii) racial, (iii) appearance-related, (iv) intellectual, (v) political, and  (vi) a generic category that contains profane words not exclusively attributed to the five specific types of harassment. A native English speaker conducted this categorization. 


\paragraph{Corpus Development and Annotation.}
We employ Twitter as the  social media data source because of its extensive public footprint. Twitter reports 313 million monthly active users that generate over 500 million tweets per day \cite{Foot20}.
Although the size of a tweet is restricted (140 characters at the time of corpus collection), once we consider a more extensive aggregation of tweets on a specific topic, mining approaches reveal valuable insights.
We utilized the first five categories of our lexicon as seed terms for collecting tweets from Twitter between December 18th, 2016 to January 10th 2017\cite{Foot20} (This date was close to the US presidential election. Then our political sub-corpus has many tweets with the subject of Trump). 
Requiring the presence of at least one lexicon item, we collected 10,000 tweets for each contextual type for a total of 50,000 tweets. As shown in Table \ref{tab:statistics}, nearly half of these tweets were annotated.
However, the mere presence of a lexicon item in a tweet does not assure that the tweet is harassing because the individuals might utilize these words with a different intention, e.g., in a friendly manner or as a quote.
Therefore, human judges annotated the corpus to discriminate harassing tweets from non-harassing tweets.
Three native English speaking annotators determined whether or not a given tweet was harassing with respect to the type of harassment content and assigned one of three labels \emph{yes}, \emph{no}, and \emph{other}. The last label indicates that the given tweet either does not belong to the current context or cannot be decided. Ultimately, we acquired $\approx$24,000 annotated tweets represented in Table \ref{tab:statistics}. Note that the annotation task was done on a per tweet basis although it can be improved using the entire conversation history.

\begin{table}[!h]
\begin{scriptsize}
\centering
\begin{tabular}{p{2cm}ccc}
 
 \toprule
\rowcolor{Gray}  \textbf{Contextual Type} & \textbf{ Annotated Tweets} & \textbf{Harassing \cmark}  & \textbf{Non-Harassing \xmark}   \\
 \midrule
 Sexual    & 3,855 & 230 &  3,619   \\
 Racial  & 4,976   &  701  &  4,275\\
 Appearance-related  & 4,828 & 678 & 4,150  \\
 Intellectual    & 4,867 & 811 & 4,056  \\
 Political  & 5,663   & 699 & 4,964 \\ \hline
 Combined  & 24,189   & 3,119 & 21,070 \\
 \bottomrule
  \end{tabular}
 \caption{\bf Annotation statistics of our categorized corpus. }
\label{tab:statistics}
\end{scriptsize}
\end{table}

\paragraph{\textbf{Agreement Rate.}}Although the annotators employed three labels, i.e., \emph{yes}, \emph{no}, and \emph{other}, the eventual corpus excluded all tweets without a consensus label of ``yes'' or ``no''. 
That is, the corpus contains only those tweets that received at least two ``yes'' or two ``no'' labels. 
Cohen's kappa coefficient \cite{cite6} measures the quality of annotation by category in Table \ref{table:agreementrate}.
The appearance-related context shows the highest agreement rate whereas political and sexual contexts have the lowest, indicating that they are more challenging to judge due to higher ambiguity.

\begin{table}[!h]
\begin{scriptsize}
\centering
\begin{tabular}{ p{2cm}c}
 \toprule
\rowcolor{Gray}  \textbf{Content Type} & \textbf{Agreement Rate}  \\
 \midrule
 Sexual &  0.70   \\ \hline
 Racial  & 0.84  \\ \hline
 Appearance-related    &  1.00      \\ \hline
 Intellectual  & 0.80     \\ \hline
 Political  & 0.69  \\ 
 \bottomrule
 \end{tabular}

 \caption{\bf Agreement rate.}
\label{table:agreementrate}
\end{scriptsize}
\end{table}

\paragraph{\textbf{Annotating Golbeck Corpus.}}The public state-of-the-art harassment-related corpus is the Golbeck corpus \cite{cite26} that only provides generic annotation, i.e., (i) harassing and (ii) non-harassing. 
This corpus contains 20,428 \textbf{non-redundant} annotated tweets of which only 5,277 are labeled as harassing.
Since we require context-aware annotations, we re-annotated the harassing tweets of Golbeck. The agreement rate (Cohen's kappa) between the two annotators is 86\%.
As shown in Table \ref{tab:statisticsGolbeck}, more than 75\% of the harassing tweets are racial. This statistic confirms Golbeck's observation. While this may be an accurate reflection of the base rate, our view is that different harassment contexts may have different consequence. An imbalanced corpus at the foundation of our research effort could result in misses of particular practical import to teenage mental health, concerning sexuality, appearance and intellect. 

\begin{table}[!h]
\begin{scriptsize}
\centering
\begin{tabular}{ p{2cm}c}
 \toprule
\rowcolor{Gray}   \textbf{Contextual Type} & \textbf{ \#of Tweets}   \\
 \midrule
 Sexual    & 380   \\ \hline
 Racial  & 4148   \\ \hline
 Appearance-related  & 145  \\ \hline
 Intellectual    & 381  \\ \hline
 Political  & 163   \\  \hline
 Non Harassing  & 41  \\
 \midrule
 Total & 5277 \\
 \bottomrule
 \end{tabular}
  \caption{\bf Statistics for the Golbeck corpus after our annotation wrt. contextual type.} 
\label{tab:statisticsGolbeck}
 \end{scriptsize}

\end{table}

\section*{LIWC Analysis for Different Types of Harassment}
\begin{figure}
\centering
\includegraphics[width=0.95\textwidth]{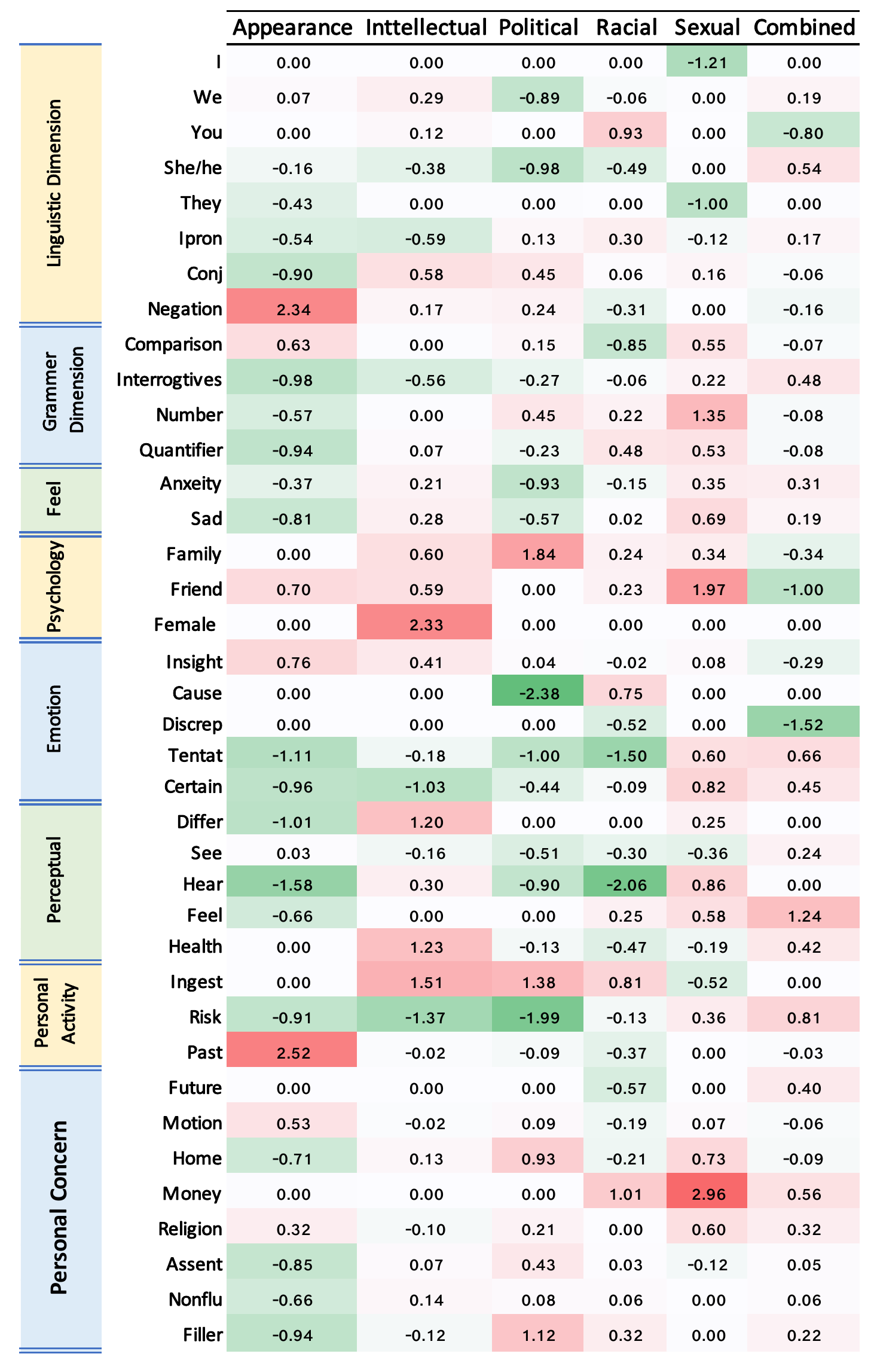}
\caption{ Significant LIWC features in comparing harassing corpus to non-harassing corpus for six categories.
The extreme red (green) color indicates the significance of a given feature in the harassing corpus (non-harassing corpus). E.g. the negation feature with the value 2.34 in the appearance harassing corpus is significantly higher than non-harassing corpus.
The white color indicates a lack of difference for a given feature when comparing two corpora.}
\label{fig:testEvaluation}
\end{figure}

Linguistic analysis of our corpus sheds light on the differences between the harassing corpus versus non-harassing corpus for each type. 
Furthermore, it provides a comparison between various types of harassment. We divided our corpus into 12 sub-corpora: (i) {\it one generic corpus\/} containing all harassing tweets regardless of their type,
called the \textbf{combined harassing corpus},
(ii) {\it one generic corpus\/} containing all non-harassing tweets irrespective of the type called the \textbf{combined non-harassing corpus}, (iii) {\it five contextual type-aware corpora\/} including only harassing tweets per type, (iv) {\it five contextual type-aware corpora\/} including only non-harassing tweets per type.  For linguistic analysis, we utilized LIWC \cite{Foot21} \cite{cite22}.
This tool tallies 96 linguistic features using a multiword lexicon for each feature. We individually analyzed each of the 12 sub-corpora using LIWC. 
An effect size , statistic estimates the magnitude of an effect (e.g., mean difference, regression coefficient, Cohen's \textit{d}, and correlation coefficient) \cite{cite38} metric was used to determine significant discriminators \cite{cite14}. 
Conventionally, a proportion (feature) $f_i$ is considered moderately discriminating when its effect size is more than $0.5$ (i.e., $|e_{f_i}|>0.5$), and is considered unhelpful if $|e_{f_i}| \approx 0$.
 The effect size for each feature is calculated as follows:
 
\begin{equation}
e_{f_i}= {\frac{\overline{experimental group}-\overline{control group}}{ std}}
\end{equation}

where, $\overline{experimental group}$ is the mean of the experimental group on the given feature $f_i$, $\overline{control group}$ is the mean of the control group wrt. the given feature $f_i$  and $std$ is the standard deviation.
For each content corpus as well as for the combined corpus, we consider the harassing corpus as the experimental group and the non-harassing corpus as the control group.
We compared the prevalence of the 96 LIWC features in the harassing corpus to their prevalence in the corresponding non-harassing corpus.
Out of the 96 original features, we removed features that were not significant in any of the contextual types and retained  38 of the most discriminating features as shown in Fig. \ref{fig:testEvaluation}. 
The extreme red (green) color represents significance (regarding effect size) of the corresponding feature in the harassing (non-harassing) corpus.
In the following, we highlight specific significant features to make three points. First, a feature is often diagnostic of the \emph{non-harassing} corpus. Second, feature significance is type dependent. The third is related to both points: a given feature, such as ``you'', can be a positive indication of harassment for one type and a negative indication of harassment for another. 
In the following, we indicate \textbf{highly significant linguistic features} derived from Fig. \ref{fig:testEvaluation} for each individual type.
Note that our corpus is already biased towards curse words because curse words are present as seeds for crawling.
Thus, our observations on discriminatory features are conditional on a ``high recall curse word-laden corpus''.

\paragraph{\textbf{Sexual Corpus}} The pronoun \emph{``I''} is prevalent in the sexually non-harassing corpus with $e=-1.2$, which is highly significant, e.g., \texttt{i'm le***an kiss}.
Furthermore, the feature \emph{``MONEY''} is prevalent in the harassing corpus with $e=2.9$.
E.g., \texttt{send free moneyb**ch hoe wont give dance hoe a*s industry bitch d*cks*** p*r* star people}.

\paragraph{\textbf{Racial Corpus}} The pronoun \emph{``YOU''} is prevalent in the harassing corpus with $e=0.9$, e.g., \texttt{Vishalp sikanda, Quideazam hahahaha u p**i  can block u cant debat u p**i I***an. } The \emph{``COMPARATIVE''} feature is prevalent in racial non-harassing corpus with $e=-0.84$, e.g., \texttt{save block p**i like po yung comment ni richard fronda} (the word `like' is an indicator of comparison in LIWC). Thus, these features can be used to discriminate 
between harassing and non-harassing tweets.


\paragraph{\textbf{Political Corpus}} The pronoun \emph{``SHE''} and \emph{``HE''}  with $e=-0.9$ and the pronoun \emph{``WE''} with $e=-0.8$ are prevalent in the non-harassing corpus, e.g., \texttt{ realdonaldtrump putin a**hat just like word can express displeasure leader god help us} (us indicates the pronoun `WE').
The \emph{``RISK''} feature is significant in non-harassing with $e=-1.9$ , e.g.,
{\texttt{f*** wrong democratic senators}}. 
The word \textit{`wrong'} represents a risk feature in LIWC dictionary. Other sample risk related words are \textit{`danger'}, \textit{`doubt'}, etc.
Furthermore, the \emph{``ANXIETY''} feature with $e=-0.92$ is significant in the non-harassing corpus.
E.g., \texttt{well i'm true dumb f*** democrat wouldn't doubt}.

\paragraph{\textbf{Appearance-related Corpus}}  \emph{ ``NEGATION''} with $e=2.3$ is prevalent in the harassing corpus  (probably because of the negative language used for referring to the body and appearance-related subjects). E.g., \texttt{Taylor swift cant shake c**el toe.}
The other significant feature in the harassing corpus is the \emph``PAST TENSE''. 
E.g., \texttt{Ugli a** didn't go run yesterday get work f*t**s.}
Furthermore, the \emph{``COMPARATIVE''} feature is prevalent  in appearance-related harassing corpus with $e=0.63$.
E.g., \texttt{ hey lardass notice your look pizza perhaps like f*** salad a**hole.}
The word \textit{`like'} indicates a comparative feature.


\paragraph{\textbf{Intellectual Corpus}} 
The \emph{ ``FEMALE REFERENCE'' } feature with $e=2.3$ is highly significant in intellectual harassing corpus (perhaps because girls are harassed more wrt. intellectual issues.)
E.g., \texttt{She is dumb f***}.

\paragraph{\textbf{Combined Corpus}}  \emph{``DISCREPANCY''} with $e=-1.5$ is prevalent in the non-harassing corpus  e.g., \texttt{ boss brought drunken sugar cook explain there alcohol just sh**face}.

\section*{Statistical Analysis of Different Types}
\label{sec:statisticalAnalysis}

\begin{figure}
\centering
\includegraphics[width=0.95\textwidth]{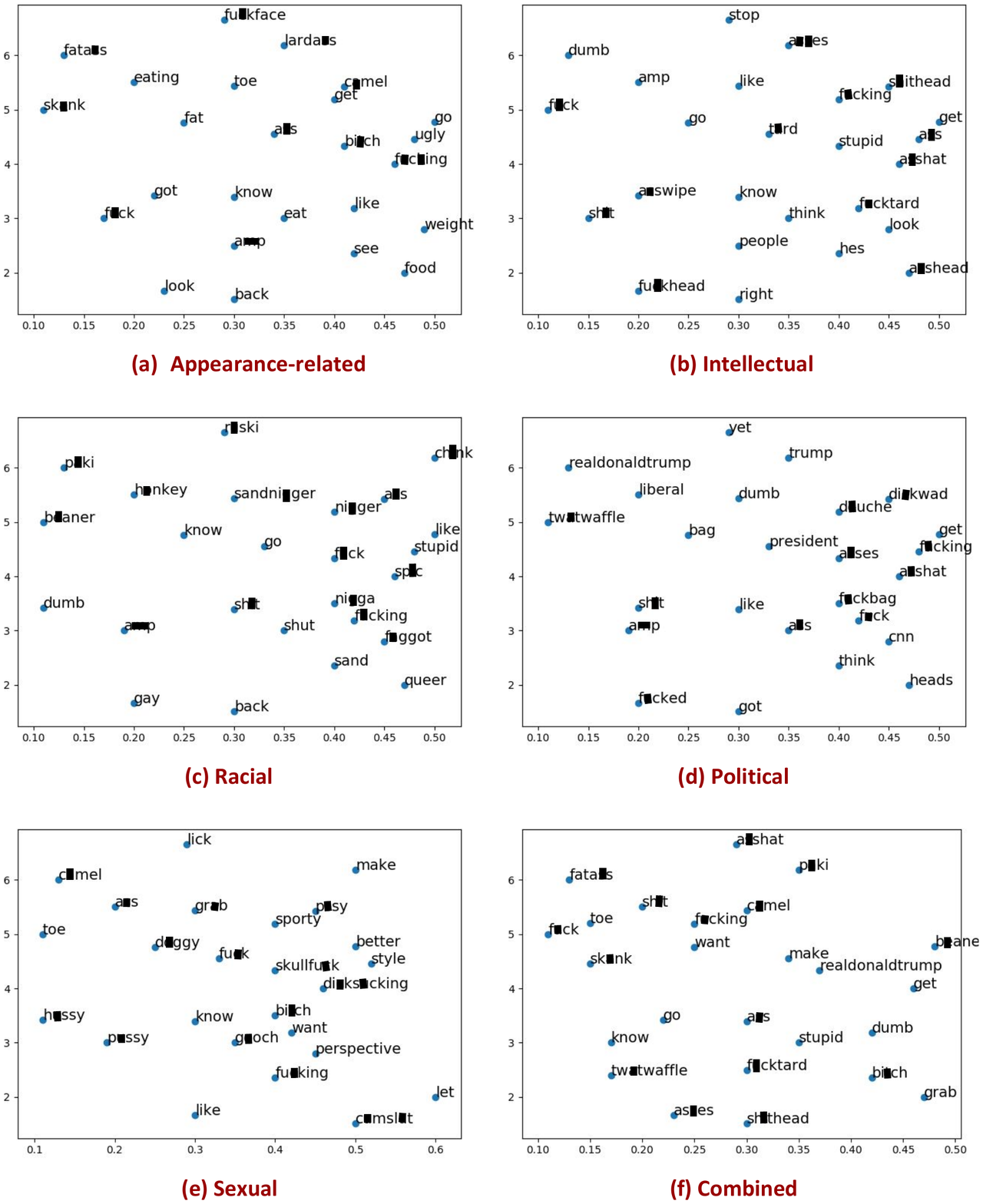}
\caption{Top-25 frequent words within each harassing corpora.}
\label{fig:harassing}
\end{figure}
\begin{figure}
\centering
\includegraphics[width=0.95\textwidth]{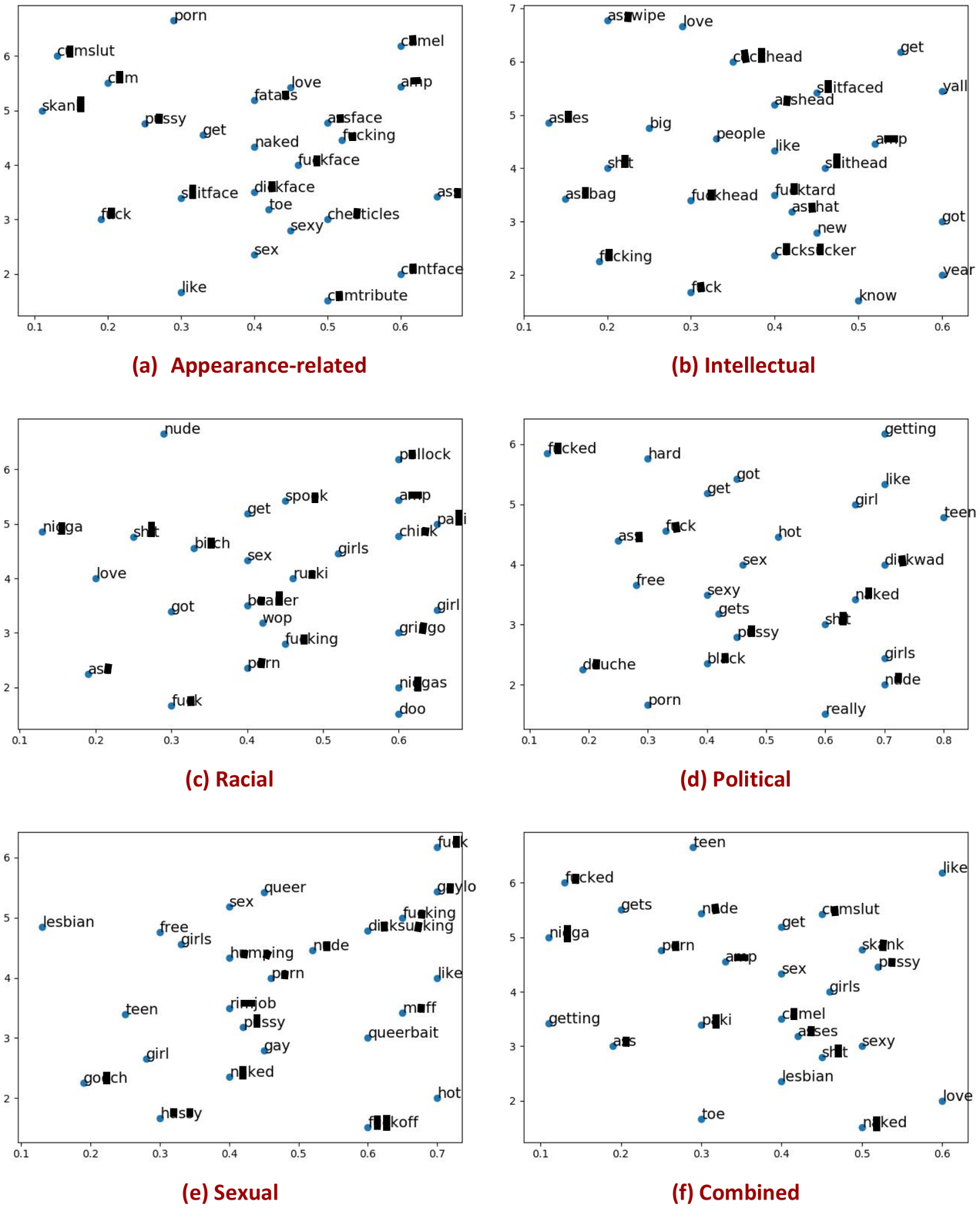}
\caption{ Top-25 frequent words within each non-harassing corpora}
\label{fig:non-harassing}
\end{figure}

We investigate the relationship between the offensive words employed in collecting our corpora and the specific lexical items in the crawled corpora. We  determine \textbf{Q1}: whether or not offensive words are observed as frequent words, \textbf{Q2}: whether or not the frequent words in harassing corpora differ from those in non-harassing corpora, and \textbf{Q3}: whether or not frequent words are type-sensitive, in other words, whether the frequent words vary with type of context. 
Fig. \ref{fig:harassing} shows the 2D visualization of the word embeddings of the top-25 most frequent words for the harassing corpora, whereas Fig. \ref{fig:non-harassing} represents a similar display for the top-25 most frequent words for the non-harassing corpora (the following section presents the details of word embedding).
The prevalence of curse words in the non-harassing corpora is comparable to the harassing corpora. This confirms that the presence of curse words is not a sufficient indicator of harassment. In the following, we mention our key observations.



\paragraph{Key Observations.}
Regarding Q1, as expected, we observed that offensive words are commonplace in both harassing and non-harassing corpora across types (cf. Fig. \ref{fig:harassing} and \ref{fig:non-harassing}).
In addition, we observed some emerging, frequent offensive words, such as ``grab'' and
``camel''that can now be added to our initial offensive lexicon \cite{Foot14}.
Furthermore,  there are frequent words that
are not necessarily offensive. E.g., consider ``look'' or ``eat'' in the
appearance-related type where they are implicitly related to the associated type, applicable to the appearance of a subject.
Regarding Q2, we observe that the frequent words in the 
harassing corpora are different from those in the non-harassing corpora.
The particular words in the harassing corpora also can be added to the initial lexicon of seed words.
The result of this analysis can be
utilized for weighting the severity of offensiveness for every
single word included in our lexicon. 

To reply quantitatively to Q3, we ran an annotation task on the top-15 most frequent words for each type of 
harassing corpus as well as the corresponding non-harassing corpus.
The description of this task is as follows: we asked the human annotators (i.e., graduate students) to determine whether or not a given frequent word is related to the associated type either explicitly or implicitly.
E.g., the words ``eat'' or ``food'' are implicitly related to appearance while they seem far from the type racial.
The results of this exercise appear in Table \ref{tab:Percentage}. 
In the harassing corpora, the percentage of relatedness of words to the associated type is higher than 67\% and in sexual and racial types, it even reaches 80\%.
This percentage fluctuates for non-harassing corpora. 
E.g., in appearance-related type, it is higher than 93\% while in racial it reaches 53\%. 
In sum, we conclude that the frequent words are mostly type-sensitive.
Moreover, the prevalence of apparently offensive language in the non-harassing corpus reinforces our claim that offensive language \textit{per se} is not necessarily harassing. 

\begin{table}[!h]
\centering
\begin{scriptsize}
\begin{tabular}{ccc}
 \hline
\rowcolor{Gray}  \textbf{Category}                            & \textbf{Type} &  \textbf{Percentage} \\ \hline
\multirow{2}{*}{\textbf{Appearance-related}} & H             & 66.6\%              \\ \cline{2-3} 
                                             & NH            & 93.3\%               \\ \hline
\multirow{2}{*}{\textbf{Intellectual}}       & H             & 73.3\%              \\ \cline{2-3} 
                                             & NH            & 73.3\%              \\ \hline
\multirow{2}{*}{\textbf{Political}}          & H             & 80\%               \\ \cline{2-3} 
                                             & NH            & 73.3\%               \\ \hline
\multirow{2}{*}{\textbf{Racial}}             & H             & 80\%                \\ \cline{2-3} 
                                             & NH            & 53.3\%              \\ \hline
\multirow{2}{*}{\textbf{Sexual}}             & H             & 80\%                  \\ \cline{2-3} 
                                             & NH            & 60\%                  \\ \bottomrule

\end{tabular}
\caption{\bf Percentage of type-dependent of top-15 frequent words within each sub-corpus. H stands for the harassing corpus and NH stands for the non-harassing corpus.}
\label{tab:Percentage}
\end{scriptsize}
\end{table}


One caveat is that the most frequent words appearing in the sub-corpus associated with each type are predominantly stop-words or curse words, as our initial seed terms are biased to an offensive lexicon.
Ignoring these words,
whose presence cuts across different types of harassment, 
revealed that the following prominent word groups are associated with 
various harassment types, shedding light on the possible features that may elicit harassment:
(i) In the appearance-related harassment corpus, 
target words such as ``eat'', ``ugly'', ``fat'', ``gym'', and ``weight'', are present. 
(ii) In the intellectual harassment corpus, 
target words such as ``dumb'', ``stupid'', ``work'', and ``head'', are present. 
(iii) In the political harassment corpus, the target words such as ``realdonaldtrump'', ``libtard'', ``dumb'', ``touch bag'', ``stupid'', and ``cnn'', are present. 
(iv) In the racial harassment corpus, 
target words such as ``m*ki'', ``n**ger'', ``b**ner'', ``ch**k'', ``muslim'', ``i**ian'', ``moron'', and  ``jew'', are present. 
(iv) In the sexual harassment corpus, 
target words such as ``hump'', ``hussy'', ``l**k'', and  ``grab'', are present. 

\section*{Predicting Different Types of Harassing Posts}
\label{sec:learning}
We aim to develop effective supervised learning methods to  detect harassing language automatically and distinguish it from non-harassing language for each contextual type. 
The state-of-the-art contains various approaches for detecting harassing content from non-harassing content but not for discriminating the type of harassment. 
We approach this gap in two ways. The first is to build individual binary classifiers that identify a particular type of harassment, e.g., a binary classifier that identifies only racial content or a binary classifier that classifies just offensive political content. The second approach uses the state-of-the-art methods to detect harassing language; after such recognition, we can employ a type-aware classifier to predict the associated type for that harassment incident.
We implemented both approaches. Initially, we trained the individual classifiers for each type.
In another approach, we built up a binary classifier that differentiates harassing content from non-harassing content regardless of their type.
Note that any classifier from the state-of-the-art can substitute for this part. Then, we built up a multi-class classifier that predicts the type of harassment incident.
The results of our experiments for both approaches reveal high accuracy.
Furthermore, to verify the effectiveness of our classifier, we apply transfer learning by running our classifier on the Golbeck corpus and assess its performance for how successfully it predicts the type of harassment.
In the following, we present the details of our experiments.

\subsection*{Transforming Tweets to Vectors}
We utilized four approaches for transforming tweets to numerical representations (i.e., vectors): (i) the conventional vectorization approach TFIDF,  (ii) word2vec, (iii) fastText and (iv) a LIWC vector.
We feed our classifiers with each of these individual vectors or a combination of them.

\paragraph{The Term Frequency and Inverse Document Frequency (TFIDF).}
We use this approach \cite{cite21} to transform each given tweet into a weighted vector T.

\paragraph{Distributional semantics (i.e., word2vec and fastText).}
Distributional semantics (so-called embedding models) \cite{word2vec1} play a vital role in many Natural Language Processing (NLP) applications. 
They capture the semantics of text units (e.g., words, characters, tweets, paragraphs or documents) from the underlying corpus and represent them in a low dimensional vector space. 
We use two major embedding models for representing each tweet. The first one is word2vec \cite{Foot22} and the second one is fastText \cite{Foot23} \cite{fasttext}.
The first one learns a dense representation at the unigram level and the second one learns at the character level.
Both of these approaches have two models, i.e., skip-gram model and CBOW model \cite{word2vec2,word2vec1} that are roughly similar. The skip-gram model (CBOW model) computes the probability of the target word $w_k$ (i.e. context word) appearing in the neighborhood of the context word $w_i$ (i.e. target word), $\mathcal{P}(w_k|w_i)$.
In this work, the vector representation of a tweet is computed as the concatenation of the vector of all tokens within the tweet.
In the rest of this paper we rely on the following notations to specify a vector.
W(S) and W(C) denote the low dimensional vector obtained respectively by the skip-gram model and CBOW model of the word2vec approach. 
F(S) and F(C) denote the low dimensional vector obtained respectively by the skip-gram model and the CBOW model of the fastText approach. 
We compiled a corpus containing 15,999,557 sentences from the Twitter and Leipzig Collection Corpora \cite{LCC} leveraging our offensive lexicon presented in \cite{ourcorpus} as the underlying seed words. Then, we trained the embedding models on this accumulated corpus using the learning parameters reported in \cite{word2vec2,word2vec1}. Our dimension size equals 300, the window size is 3, and the minimum count equals 10.

\paragraph{LIWC Vector.} The vector obtained by running the LIWC tool is denoted by L.

\subsection*{Evaluation of the Harassment Classifiers}
\label{sec:evallearning}

\paragraph{Preparing training datasets.}
As the number of harassing tweets is not equal to the number of non-harassing ones in our corpus -- in fact, it varies for each type -- we prepared balanced datasets for training the classifier. 
We prepared five type-aware training data sets using an under-sampling approach taking all of the harassing tweets with an equal number of non-harassing (randomly sampled).
Also, we prepared a combined training data set considering all of the harassing tweets regardless of their type and an equal number of non-harassing tweets.
Table \ref{tab:StatisticOfDataset} shows the size of the training data sets for each type.
Each data set contains an equal number of harassing tweets versus non-harassing tweets. 
Later, we employ the remaining tweets to test the robustness of the classifiers against unseen data.

\begin{table}[!h]
\centering
\begin{scriptsize}
\begin{tabular}{cc}
\toprule
\rowcolor{Gray}  \textbf{Category}           & \textbf{Number of tweets} \\ \midrule
\textbf{Appearance-related} & 1,344                          \\ \hline
\textbf{Intellectual}       & 1,622                          \\ \hline
\textbf{Political}          & 1,397                          \\ \hline
\textbf{Racial}             & 1,401                          \\ \hline
\textbf{Sexual}             & 461                           \\ \midrule
\textbf{Combined}             & 6,225                         \\ \bottomrule
\end{tabular}
\end{scriptsize}
\caption{\bf Size of the training datasets for each type.  }
\label{tab:StatisticOfDataset}
\end{table}

\begin{figure}[hptp]
\centering
\includegraphics[width=\textwidth]{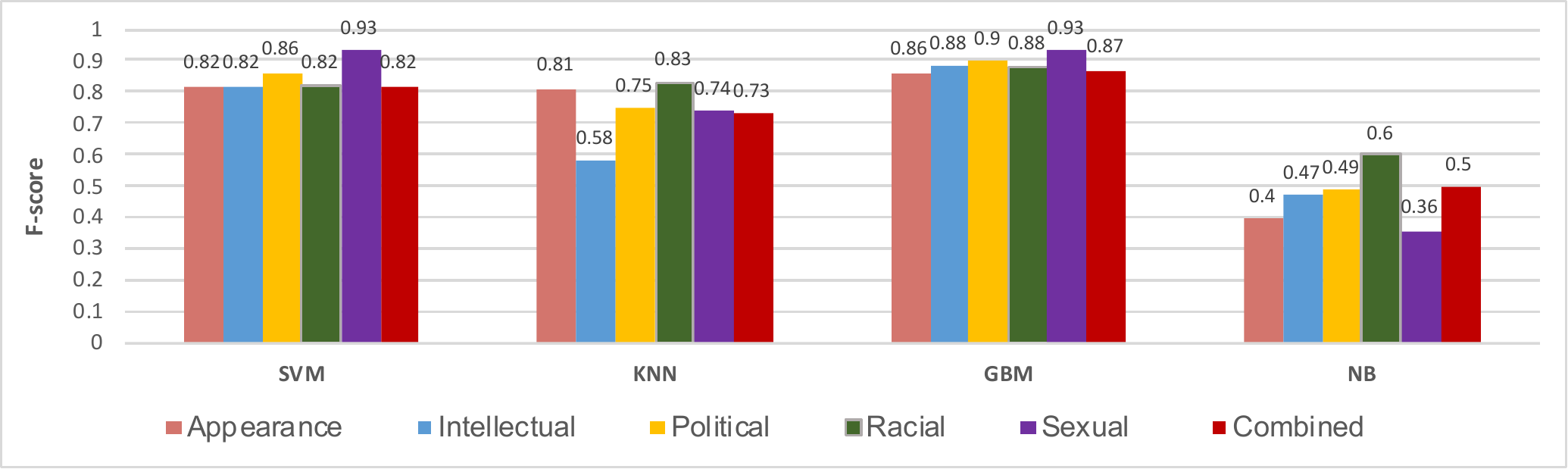}
\caption{ Comparative Study of the F-score from four major classifiers i.e., SVM stands for Support Vector Machine, KNN= K-Nearest Neighbor, GBM= Gradient Boosting Machine, NB= Naive Bayes, NN= Nueral Network)}
\label{fig:binaryClassifier-comparison}
\end{figure}

\paragraph{Training Binary Classifiers.} In our experimental study, we trained four types of classifiers, using (i) \emph{Support Vector Machine} (SVM) \cite{cite23}, (ii) \emph{K-Nearest Neighbors } (KNN) \cite{cite23}, (iii) \emph{Gradient Boosting Machine} (GBM) \cite{cite37}, and (iv) \emph{Naive Bayes} (NB) \cite{cite23}.
We rely on the following settings for the GBM classifier: the learning rate is 0.1, loss function is logistic regression, the number of trees is 100, sub-sample is 1.0, the criteria function is Friedman MSE, the minimum sample is 2, the minimum number of samples required to be at a leaf node is 1, and the maximum depth of the individual regression estimators is 3.
We ran 10-fold cross-validation with re-sampling and iteration strategies (repeated five times).
Figure \ref{fig:binaryClassifier-comparison} shows the performance of the classifiers based on an F-score measure using a TFIDF vector.
Generally, the results of the NB classifier in all of the cases were inferior whereas the GBM classifier outperforms others in the majority of settings except for a few instances comparable to the SVM classifier. 
Thus, in the following experiments, we rely on the GBM classifier.

\begin{figure}[hptb]
\centering
\includegraphics[width=0.8\textwidth]{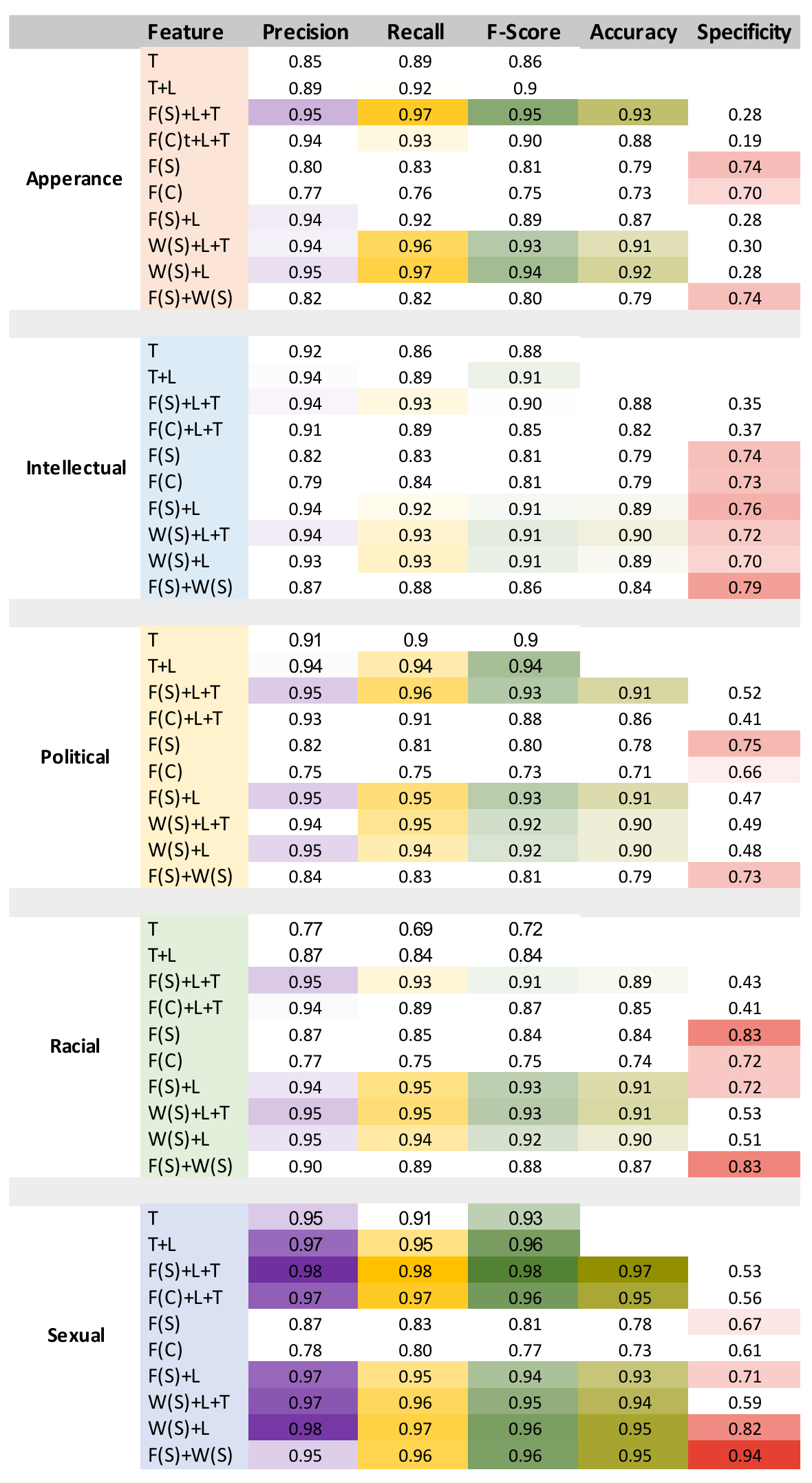}
\caption{ Comparative Study of the various feature settings on the performance of the GBM classifier using measures such as precision, recall, F-score, accuracy, and specificity. The extreme colors, i.e., purple, yellow, green, olive, and pink show the higher values versus the white color that shows a lower value.}
\label{fig:featureengineering}
\end{figure}

\paragraph{Feature Engineering.} 
To gain insight over the effectiveness of various features, we feed the GBM classifier with various feature settings. The fine-grained results of our experiment are listed in Figure \ref{fig:featureengineering}.
We employed a various combination of vectors, for example, F(S)+W(S) means the input features were the skip-gram models of fastText and word2vec.
In addition to the typical precision, recall and F-score measures, we provide specificity (True negative rates) and accuracy rates.
We offer the following observations: 
(i) The tweet representation using F(S)+W(S) vector is the most effective input representation as it provides high and balanced rates for all measures including precision, recall, F-score and specificity.
Note that in multiple settings such as F(S)+L+T, precision, recall, and f-score are high whereas specificity is low meaning that the classifier is biased towards one of the classes and does not perform reasonably on both classes.
(ii) In the settings for which the LIWC vector L is included, typically the specificity rate is low. This probably means L vector does not provide a discriminative representation for the classifier.
(iii) Generally learning the representation of tweets using the fastText approach either with skip-gram or CBOW shows high performance. This might come from the fact that encoding tweets at the character level is more effective for detecting harassment.
(iv) The sexual type resulted in the classifier with the highest accuracy (with F-score 96\% and specificity 94\%), racial and intellectual are in the next positions (respectively with F-score 88\%, 86\% and specificity 83\%, 79\%).  

\paragraph{Binary Classifier for Harassment Detection.}
We also trained a binary classifier on our combined corpus where it can differentiate the harassing language from non-harassing regardless of the contextual type.
In situations that the type of harassment does not play a role, or type detection must occur after the harassment detection, using such a generic classifier is necessary.
Table \ref{tab:Binary-CombineClassifier} shows the detailed results of this classifier in various settings of input features.
Generally, the vector of FastText F shows an effective role, especially when it is coupled with the W vector; the specificity score reaches its optimum.

\begin{table}[h!]
\centering
\begin{scriptsize}
\begin{tabular}{cccccc}
\toprule
\rowcolor{Gray} 
\multicolumn{1}{l}{\textbf{Feature}} & \multicolumn{1}{l}{\textbf{Precision}} & \multicolumn{1}{l}{\textbf{Recall}} & \multicolumn{1}{l}{\textbf{F-Score}} & \multicolumn{1}{l}{\textbf{Accuracy}} & \multicolumn{1}{l}{\textbf{Specificity}} \\ \midrule
T                                                            & 0.84                                                           & 0.81                                                        & 0.82                                                         &                                                               &                                                                  \\ \hline
T+L                                                          & 0.9                                                            & 0.87                                                        & 0.88                                                         &                                                               &                                                                  \\ \hline
F(S)+L+T                                                     & 0.94                                                           & 0.92                                                        & 0.90                                                         & 0.88                                                          & 0.37                                                             \\ \hline
F(C)+L+T                                                     & 0.94                                                           & 0.88                                                        & 0.86                                                         & 0.84                                                          & 0.44                                                             \\ \hline
F(S)                                                         & 0.83                                                           & 0.83                                                        & 0.82                                                         & 0.80                                                          & 0.75                                                             \\ \hline
F(C)                                                         & 0.78                                                           & 0.76                                                        & 0.76                                                         & 0.75                                                          & 0.73                                                             \\ \hline
F(S)+L                                                       & 0.94                                                           & 0.95                                                        & 0.93                                                         & 0.91                                                          & 0.69                                                             \\ \hline
W(S)+L+T                                                     & 0.94                                                           & 0.93                                                        & 0.91                                                         & 0.89                                                          & 0.70                                                             \\ \hline
W(S)+L                                                       & 0.93                                                           & 0.94                                                        & 0.92                                                         & 0.90                                                          & 0.74                                                             \\ \hline
F(S)+W(S)                                                    & 0.90                                                           & 0.89                                                        & 0.88                                                         & 0.87                                                          & 0.83                                                             \\ \bottomrule
\end{tabular}
\end{scriptsize}
\caption{\bf Performance of the GBM binary classifier on  the combined corpus. }
\label{tab:Binary-CombineClassifier}
\end{table}

\paragraph{Type Prediction using a Multi-Class Classifier.}
Apart from building binary classifiers for predicting types, we trained a multi-class classifier to predict the type of harassment incidents.
We trained several multi-class classifiers, among them the GBM classifier outperformed others. Herein we report the result for GBM classifier only. 
We used W(S)+F(S) vectorization approach as the input feature.
Then, we trained this classifier on a corpus containing all of the sub-corpora from the previous step. This corpus has samples with six various labels where five labels indicate a particular type of harassment and the last label indicates ``non-harassing'' implying there is no harassing language.
Table \ref{tab:multi-classClassification} shows the details of the evaluation on the performance of this classifier where the micro F-score is 0.92 and the macro F-score is 0.82.
Note that in the macro-level, we calculate the performances of each class and then average whereas, in the micro-level, we calculate the performance for all classes, as computing contingency table and then evaluate precision/recall and F-score \cite{jurafsky2014speech}.
Digging into fine-grained efficiency shows that the accuracy across various classes holds similar behaviors except for a decrease in the precision and recall of the sexual type. As we will discuss in error analysis below, this type is prone to mis-classification with the other types particularly the racial type.
However, comparing the performance of multi-class classifier and binary classifiers shows that the multi-class classifier mostly outperforms the binary classifiers by as much as $\approx 10\%$.
Note that the accuracy of our classifier will improve on a generic tweet corpus because our current 
corpus has been crawled using curse words with a significantly higher proportion of harassing tweets 
compared to that in a generic tweets corpus, which is predominantly non-harassing and devoid of curse words.
On the downside, it will miss harassment conveyed through ``clean'' words.
However, to demonstrate the effectiveness of the current version of this classifier, in the next step we apply it on an unseen corpus to predict the type of harassment incident.


\begin{table}[!h]
\centering
\begin{scriptsize}
\begin{tabular}{lcclc}
\toprule
\rowcolor{Gray}  \textbf{Category}          & \textbf{Precision} & \textbf{Recall} &  \textbf{F-score} \\ \midrule
\textbf{Appearance-related} & 0.84 & 0.85 & 0.84                           \\ \hline
\textbf{Intellectual}  &   0.87  & 0.85 & 0.86                          \\ \hline
\textbf{Political}    &  0.81    & 0.84 &  0.83                         \\ \hline
\textbf{Racial}    &     0.82    & 0.83  & 0.82                       \\ \hline
\textbf{Sexual}    &    0.58   & 0.62 &  0.60                        \\ \hline
\textbf{Nonharassing}    &     0.98   & 0.97 &  0.98 
\\ \bottomrule
\toprule
\textbf{Micro Precision} & 0.92      & \textbf{Macro Precision}    &   0.82      \\ \hline
\textbf{Micro Recall}  &   0.92       & \textbf{Macro Recall}    &  0.83    \\ \hline
\textbf{Micro F-score}    &  0.92     & \textbf{Macro F-score}    &     0.82        \\ 
\bottomrule
\end{tabular}
\end{scriptsize}
\caption{\bf Performance of our multi-class classifier for predicting type of harassment incident. }
\label{tab:multi-classClassification}
\end{table}

\paragraph{Comparison to the state-of-the-art.} Since this work was the first to introduce contextual type for harassment, comparison to the state-of-the-art that relies only on two or three variants of harassment, is unfair. However, to verify the effectiveness of our type-oriented multi-class classifier, we tested it on the harassing tweets from the Golbeck corpus (an external corpus unseen to our classifier) that is a publicly available state-of-the-art harassment-related corpus \cite{cite2}.
This corpus contains 20,428 annotated tweets of which only 5,277 are labeled as harassing. It does not distinguish the nature of the harassment. 
In Table \ref{tab:statisticsGolbeck}, we represented our annotations for the harassing tweets of the Golbeck corpus with respect to our types using human judges which yielded in an agreement rate of 86\%.
The proportion of harassing tweets per type is represented in the last column of Table \ref{tab:GolbeckClassification}.
We ran our type-aware multi-class classifier (GMB classifier) to predict the associated type of harassing tweets on Golbeck corpus.
Table \ref{tab:GolbeckClassification} shows the precision, recall and F-score for each type.
We observe an F-score of more than 94\% for all types except for the type appearance. In the case of the racial type, the F-score reaches 98\%.
This high performance exceeds the state-of-the-art where they are mostly concerned about detecting the general harassing language (the reported accuracy ranges between 70\% and 85\%) \cite{cite27,cite28, cite43, cite44}.
In addition, it shows robustness with unseen data.
Note that the racial type is dominant in the Golbeck corpus.
We also ran our classifier on a portion of 5,000 non-harassing tweets from Golbeck corpus, which resulted in the F-score > 98\% (cf. Table \ref{tab:GolbeckClassification}). 
The last three rows of Table \ref{tab:GolbeckClassification} show micro and macro precision, recall, and F-score. The closeness of the micro and macro measures shows that the classifier is not biased towards a dominant class.

\begin{table}[!h]
\centering
\begin{scriptsize}
\begin{tabular}{lcclc}
\toprule
\rowcolor{Gray}  \textbf{Category}          & \textbf{Precision} & \textbf{Recall} &  \textbf{F-score} & \textbf{Proportion Rate}\\ \midrule
\textbf{Appearance-related} & 0.74 & 0.63 & 0.68 &2.7\%                          \\ \hline
\textbf{Intellectual}  &   0.91  & 0.92 & 0.91 &7.2\%                         \\ \hline
\textbf{Political}    &  0.90    & 0.95 &  0.92&3.0\%                         \\ \hline
\textbf{Racial}    &     0.99    & 0.97  & 0.98 &78.6\%                       \\ \hline
\textbf{Sexual}    &     0.94   & 0.96 & 	0.95  & 7.2\%
\\ \hline
\textbf{Nonharassing}    &     0.99   & 0.98 &  0.98 & \-\-
\\ \bottomrule
\toprule
\textbf{Micro Precision} & 0.97      && \textbf{Macro Precision}    &   0.91      \\ \hline
\textbf{Micro Recall}  &   0.97       && \textbf{Macro Recall}    &  0.90    \\ \hline
\textbf{Micro F-score}    &  0.97     && \textbf{Macro F-score}    &     0.91        \\ 
\bottomrule
\end{tabular}
\end{scriptsize}
\caption{\bf Performance of our classifier for predicting tweets for Golbeck corpus.  }
\label{tab:GolbeckClassification}
\end{table}


\paragraph{Error Analysis.} To make sense of classifier errors, we examined a couple of tweets classified as sexual.
E.g., for \texttt{@usr you deserved to be raped by a thousand Muslims in your c**t a**hole }, our classifier classified that as sexual harassment and not racial because of the word `rape'.
Similarly, the tweet \texttt{@usr @usr lol it's not against women. It's against  f***ing feminist c***s like you.  \#feminazi \#womenagainstfeminism} was classified as sexual.
Such cases are ambiguous because even manual annotation is highly subjective. 
In other words, categorizing harassment is highly subjective and the boundary between types is not rigid.
In majority of the overlapping cases (racial and sexual),  the tweets were classified as sexual rather than racial.
We also analyzed errors in political tweets and concluded that harassment signal can be: (i) implicit, e.g., \texttt{John Boehner blames Democrats for \#shutdown. He better stop drinking cuz a few more drinks and he starts blaming the J*ws f}, (ii) ambiguous \texttt{??? You're a wh*** to the telecom industry, i hope your constituents vote you out.}, (iii) unreliable, e.g., \texttt{It's going to be a republican government in the US next term. Democrats can kiss their presidency bid goodbye. Let the J*ws rule!}, (iv) poorly captured through annotation, e.g., the tweet \texttt{@TrueNugget @FeministPeriod @OregonState Man college is becoming more and more a mistake.}  in the Golbeck corpus. 
Our classifier misses them as they are weak cases of harassment.

\section*{Ethics}
We anonymized data. The study was approved by the Wright State University Institutional Review Board entitling "Student Use of Social Media" as protocol number IRB \#: 06251.

\section*{Conclusion and Future Plans}

In this paper, we introduced five contextual types for harassment, namely,
(i) sexual, (ii) racial, (iii) intellectual, (iv) appearance-related and (v) political.
We presented experiments with a type-aware tweets corpus to analyze, learn, and understand harassing language for each type.
Our contribution lies in providing a systematic and comparative approach to assessing harassing language from linguistic and statistical perspectives. 
Furthermore, we built type-specific classifiers, and the results of our experiments show the importance of considering the contextual type 
for identifying and analyzing harassment on social media.

In general, a single tweet identified as ``harassing'' may not provoke the same intense negative feeling that we associate with that word in the real-world scenario. However, in practice,  ``conversational'' exchanges
containing a sequence of such tweets can rise to the level of harassment causing mental and psychological anguish,
and fear of physical harm. Nevertheless, our current Twitter dataset is limited to annotating single tweets in isolation for harassment. Furthermore, the reliable assessment of the type of harassment is a difficult problem because it requires significant knowledge of current events and common-sense.
We plan to extend this work by learning the language of harassers as well as victims, and further study the contribution of non-verbal cues (i.e., conversational features, network features, and community features) for identifying online harassment activities, particularly on social media.

\section*{Acknowledgment}
We acknowledge support from the National Science
Foundation (NSF) award CNS 1513721: Context-Aware Harassment Detection on Social Media.
Any opinions, findings, and conclusions, recommendations expressed in this material
are those of the author(s) and do not necessarily reflect the views of the NSF.

\bibliographystyle{plain}
\bibliography{reference} 

\end{document}